\documentclass[preprint,12pt]{elsarticle}

\usepackage{amssymb}
\usepackage[bookmarks=false]{hyperref}
\usepackage{tabularx}
\usepackage{multirow}
\usepackage{svg}
\usepackage{lineno}
\journal{Journal}

\begin{document}

\begin{frontmatter}

%% Title, authors and addresses

%% use the tnoteref command within \title for footnotes;
%% use the tnotetext command for theassociated footnote;
%% use the fnref command within \author or \address for footnotes;
%% use the fntext command for theassociated footnote;
%% use the corref command within \author for corresponding author footnotes;
%% use the cortext command for theassociated footnote;
%% use the ead command for the email address,
%% and the form \ead[url] for the home page:
%% \title{Title\tnoteref{label1}}
%% \tnotetext[label1]{}
%% \author{Name\corref{cor1}\fnref{label2}}
%% \ead{email address}
%% \ead[url]{home page}
%% \fntext[label2]{}
%% \cortext[cor1]{}
%% \affiliation{organization={},
%%             addressline={},
%%             city={},
%%             postcode={},
%%             state={},
%%             country={}}
%% \fntext[label3]{}

\title{Plant Disease Recognition Datasets in the Age of Deep Learning: Challenges and Opportunities}
\date{}

%% use optional labels to link authors explicitly to addresses:
%% \author[label1,label2]{}
%% \affiliation[label1]{organization={},
%%             addressline={},
%%             city={},
%%             postcode={},
%%             state={},
%%             country={}}
%%
%% \affiliation[label2]{organization={},
%%             addressline={},
%%             city={},
%%             postcode={},
%%             state={},
%%             country={}}

\author[1]{Mingle Xu}
\author[1]{Ji Eun Park}
\author[1]{Jaehwan Lee}
\author[2]{Jucheng Yang \corref{cor}}
\ead{jcyang@tust.edu.cn}
\author[3]{Sook Yoon \corref{cor}} 
\ead{syoon@mokpo.ac.kr}

\cortext[cor]{Corresponding author}

\affiliation[1]{organization={Department of Electronic Engineering, Core Research Institute of Intelligent Robots, Jeonbuk National University},
city={Jeonju},
country={South Korea}
}
\affiliation[2]{organization={College of Artificial Intelligence, Tianjin University of Science and Technology},
city={Tianjin},
country={China}
}
\affiliation[3]{organization={Department of Computer Engineering, Mokpo National University},
city={Muan},
country={South Korea}
}

\begin{abstract}
%% Text of abstract
Plant disease recognition has witnessed a significant improvement with deep learning in recent years. Although plant disease datasets are essential and many relevant datasets are public available, two fundamental questions exist. First, how to differentiate datasets and further choose suitable public datasets for specific applications? Second, what kinds of characteristics of datasets are desired to achieve promising performance in real-world applications? To address the questions, this study explicitly propose an informative taxonomy to describe potential plant disease datasets. We further provide several directions for future, such as creating challenge-oriented datasets and the ultimate objective deploying deep learning in real-world applications with satisfactory performance. In addition, existing related public RGB image datasets are summarized. We believe that this study will contributing making better datasets and that this study will contribute beyond plant disease recognition such as plant species recognition. To facilitate the community, our project is public \href{https://github.com/xml94/PPDRD}{available} with the information of relevant public datasets.
\end{abstract}

%%Graphical abstract
% \begin{graphicalabstract}
% \includegraphics{grabs}
% \end{graphicalabstract}

%%Research highlights
% \begin{highlights}
% \item Research highlight 1
% \item Research highlight 2
% \end{highlights}

\begin{keyword}
%% keywords here, in the form: keyword \sep keyword
plant disease dataset \sep deep learning \sep smart agriculture \sep precision agriculture
%% PACS codes here, in the form: \PACS code \sep code
% \PACS 0000 \sep 1111
%% MSC codes here, in the form: \MSC code \sep code
%% or \MSC[2008] code \sep code (2000 is the default)
% \MSC 0000 \sep 1111
\end{keyword}

\end{frontmatter}

% \linenumbers

%% main text
\section{Introduction}
% History of plant disease recognition, the advantages of deep learning.
% dataset difference between traditional machine learning and deep learning.
% the role of plant disease dataset which does not receive enough attention.
% what is the problem towards datasets.
% why disease dataset is hard to get, compared to ImageNet.
% Our contribution

To have enough food is a basic requirement of human beings. However, more than 600 million people worldwide are estimated to be exposed to hunger in 2030 according to the \href{https://sdgs.un.org/goals/goal2}{United Nations}. However, many things threaten the availability of food and plant disease is one of the most essential. It is estimated that \$220 billions loss because of plant disease according to the \href{https://www.fao.org/plant-health-2020/home/en/}{Food and Agriculture Organization of the United Nations}. It is therefore eager to mitigate them and recognizing plant diseases is a fundamental mission. However, a traditional way is that human experts have to go farm to see the plants and then make decisions. This paradigm is expensive and noisy because training experts need time and many factors play a role to human to make decisions such as mood and lasting time to work \cite{kahneman2021noise}.

Fortunately, deep learning has been showcased the potential to recognize plant diseases automatically in recent years \cite{singh2018deep, liujun2021plant, xu2022transfer, thakur2022trends, xu2023enhanced, salman2023crop}. To achieve a decent performance for deep learning models, dataset is one of the most essential considerations \cite{krishna2017visual, cui2022stable, wright2022high, xu2023comprehensive}. High-quality training datasets are expected to achieve decent test performance and superior generalization capacity. However, plant disease recognition datasets have received relatively less attention in recent years. We argue that it is worthwhile with the following reasons. First, there is no high quality widely used benchmark, which seriously hinders the further development. A common observation is that current deep learning methods tends to suffer in the real-world applications, especially in generalization in general computer vision tasks \cite{arjovsky2019invariant, bengio2021deep, cui2022stable, corso2023holistic}, as well as in the recognition of plant diseases \cite{wu2023laboratory, guth2023lab, xu2023embracing}. The field of plant disease recognition desire those datasets in high-quality and close to real-world applications. Second, this kind of dataset is much difficult to collect as it needs essential understandings in both deep learning and agricultural field. Compared to generic benchmarks in computer vision such as ImageNet \cite{deng2009imagenet} and COCO \cite{lin2014microsoft} that are related to daily-available objects, strong domain knowledge about agriculture and plant is elusively required to create a plant disease recognition dataset. Simultaneously, knowledge about deep learning should be involved, such as the challenges with current deep learning methods and the strategies to annotate the data. For example, data annotation should be compatible with the application objective and deep learning methods, as detailed in Section \ref{annotation}.

To address such issues, this paper aims to bridge agriculture and deep learning, which will enhance the understanding to make superior related datasets and further facilitate the deployment of deep learning in real-world tasks of plant disease recognition. Our ambitious objective it to deploy deep learning models in real-world applications effectively, efficiently,  and robustly. With such goal, this study has the following main contributions:
\begin{itemize}
    \item It proposes an informative taxonomy for plant disease recognition datasets.
    \item It summarizes current public plant disease datasets.
    \item It presents the future directions to create plant disease datasets with additional discussions.
\end{itemize}

\section{Taxonomy}
As shown in Table \ref{tab:taxonomy}, this section aims to provide a taxonomy to describe and distinguish datasets to recognize plant diseases. We emphasize that diverse datasets have their disadvantages and advantages, and thus should be considered from different perspectives, which inspires this section. We hope the proposed taxonomy will enhance the understanding for the community from making the objective of real-world applications, collecting suitable datasets, to deploying compatible deep learning methods.

\begin{table}[htbp]
    \centering
    \caption{Taxonomy of datasets to recognize plant diseases using deep learning.}
    \label{tab:taxonomy}
    \small
    \begin{tabularx}{\textwidth}{|l|l|X|}
       \hline 
       \multicolumn{2}{|l|}{Application objective} & It can be considered from interest, including types of plant and organ, and recognition levels, including classification, localization, and quantitation. \\
       \hline
       \multicolumn{2}{|l|}{Input modality} & It covers optical image, video, text, audio, and their combinations. \\
       \hline 
       \multirow{2}{*}{Image acquirement} & Optical sensor & Type of sensors to get images, including the cameras of RGB, hyper-spectral, multi-spectral, thermal, depth image. \\
       \cline{2-3}
            & Platform  & Place or device to put the optical sensors, including human hand, robot arm, UAV, aircraft, and satellite. \\
       \hline 
       \multicolumn{2}{|l|}{Image variation} & Change and visual variation of images within a class, such as background, illumination, and scale. The image belonging to a class in a dataset may be with enormous or few image variation. \\
       \hline 
       \multicolumn{2}{|l|}{Dataset splitting} & Strategies to split a collected dataset into training, test, and validation datasets, including random, spatial, and temporal. \\
       \hline
       \multirow{3}{*}{Annotation} & Existence & Datasets can be categorized into fully, partly, and no annotated if every image, part of images, and non images are annotated, respectively. \\
            \cline{2-3}
            & Correctness & Strategy to make sure that annotations from human experts are correct. In general, annotations are with bias and noise and voting is an effective yet expensive strategy to reduce them if experts provide annotations individually. \\
            \cline{2-3}
            & Level & Annotation level, including image, instance, and pixel level where annotations are given for a holistic image, every instance of disease, and every pixel. \\
       \hline
       % \multicolumn{2}{|l|}{Challenge} & Challenges to achieve competitive performance with the specific datasets using deep learning. \\
       % \hline
    \end{tabularx}
\end{table}

\subsection{Application objective}

In terms of plant disease recognition, different applications may have specific \emph{interests}, such as the type of plants and organs. For example, some applications focus on one specific crop such as tomatoes \cite{fuentes2017robust, xu2022style} and apples \cite{thapa2020plant} whereas others may consider multiple crops \cite{hughes2015open, liu2021plant}. Similarly, diseases exist in different organs, such as leaves, fruits, and stems. 

Moreover, applications require different recognition levels. When diseases appear, one may wonder what it is, referring to classification. Sometimes, multiple diseases may happen and localization is thus beneficial. Specifically, one plant may be with more than one unhealthy symptom where the locations give more precise information, as well as one leaf. Furthermore, some decisions and remedies can be adopted based on their magnitudes, termed quantitation, such as the number of infected leaves and the severity in an unhealthy leaf. To some extent, the complexities of the mentioned analysis gradually improves. Fortunately, these analysis can be implemented by choosing appropriate methods from deep learning, including image classification, object detection, and segmentation \cite{xu2023embracing}.

\subsection{Input modality}

To recognize plant diseases, human experts use multiple senses such as vision and smell. In addition, knowledge from other experts and their own experience also benefits. In terms of machines equipped with deep learning, similar scenarios exist. Optical images, a kind of vision, are one of the most fundamental modalities of information to recognizing plant diseases. They can be obtained with different devices and with multiple sub-categorizes, as described in the next subsection. Videos and time-series images provide extra information than the images alone. To be more specific, videos may capture the visual patterns of plant diseases from different perspectives and distances which can be taken as accumulated observations. In a similar spirit, time-series images resemble the actions of human experts who investigate the transformation of plant diseases in a row of days to make decisions. Besides, texts are also beneficial with semantic information in nature because they are created by human beings. For example, it depict the characteristics of plant diseases such as color and their changes in temporal. Texts can describe images such as locations of diseases and their magnitudes of severity \cite{fuentes2019deep, wang2022plant, cao2023cucumber}. Furthermore, texts can be replaced with audio to give human knowledge. More modalities are possible and encouraged such as smell and novel ones because of new types of sensors can also be employed in the future \cite{zhang2023wearable}.

\subsection{Image acquirement}

\begin{figure}[htbp]
    \centering
    % \includesvg[inkscapelatex=false, width=0.6\textwidth]{platform_sensors.svg}
    \includegraphics[width=0.6\linewidth]{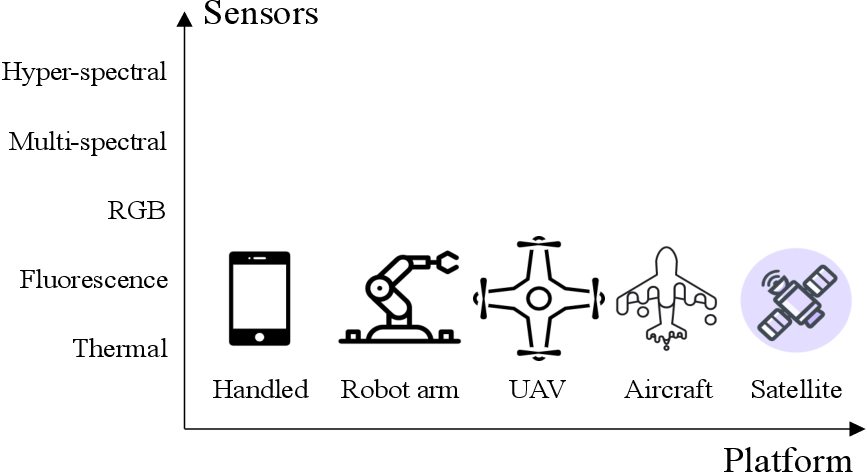}
    \caption{Optical sensors and platform to take images for plant disease, adapted from \cite{xu2023enhanced} and inspired by \cite{mahlein2016plant, oerke2014proximal}. Extra types of sensor and platform are possible and encouraged. Please refer the advantages and disadvantages of the sensors and platform to \cite{mahlein2016plant, oerke2014proximal}.}
    \label{fig:platform}
\end{figure}

Although heterogeneous modalities of input exist to recognize plant diseases as mentioned above, optical images are the most widely one. This subsection aims to probe the ways to get them since there are multiple types for optical image beneficial for diverse cases \cite{oerke2014proximal, mahlein2016plant}. As shown in Figure \ref{fig:platform}, optical image acquirement is grouped into two factors, sensors to take images and platform to hold the sensors.

The widely used type of optical sensor is the RGB (red, green, and blue) camera that captures a range of visible wave length. Human understand RGB images very well. One of the reason of the popularity of RGB images is the great availability resulting from the relative cheap mobile phones. Simultaneously, those phones handled by human beings produce enormous RGB images. Imagine the scenario that any person with a phone can take pictures if they are interested in the abnormal plants. Besides, The resolution evolves larger significantly with clear details. Except for human, RGB cameras can be fixed to monitor the growth of plants. RGB cameras placed in robot arms that can move automatically would be an efficient way to free human. We argue that this type is superior to the plants in greenhouses. In spite of its super convenience of RGB images, extra information is also contributing. For example, thermal sensors are light-free and thus can be employed in the night when RGB cameras fail to work. Fluorescence is another possible type although, to the best of our knowledge, there is no related dataset.

The above mentioned sensors take images with a certain range of wavelength. In contrast, multi- and hyper-spectral sensors can also record images with multiple and many levels of wavelength, resulting in images with many channels. In general, many vegetation indexes can be obtained with the two sensors \cite{adao2017hyperspectral, lu2020recent, lu2019comparing, wan2022hyperspectral}. One of the main advantages is that they are potential to capture images for a large area beyond single leaf and plant \cite{oerke2014proximal, mahlein2016plant, xu2023enhanced}. Relatively, those two sensors are generally put into UAVs (unmanned aerial vehicle) and aircraft to surveil many plants. However, their disadvantages are the non-trivial computations resulting from the many channels and inconvenience to use.

\subsection{Image variation}

In the age of traditional machine learning, engineers and researchers carefully consider data collection and thus the collected data are relatively small yet informative \cite{wright2022high}. This situation has changed in the era of deep learning where the datasets become much larger yet non-informative. Sometimes, datasets are collected without any specific objective in advance \cite{wright2022high}. Analyzing these datasets is therefore essential and variation is arguable one of the most important for image-based datasets \cite{fuentes2017spectral, PlantDoc, xu2023comprehensive, xu2023embracing, xu2023enhanced, wu2023laboratory}. The inspiration is to achieve decent generalization performance and a basic assumption of machine learning and deep learning is that the training and test datasets are in an identical and independent distribution (i.i.d) \cite{vapnik1991principles}. However, this assumption is not hold in many real-world application. Hence, we contend that understanding variation within the collected dataset is beneficial to making robust applications and this study focuses on the RGB image variation because of its prevalence in recent years. 

Officially, image variations consist of inter-class, the diversity between two classes, and intra-class, the diversity within one class \cite{xu2023embracing}. One of the basic assumption to distinguish plant diseases is that different diseases have different visual patterns even if they are similar \cite{xu2023embracing}; otherwise, methods belonging to pattern recognition and classification fail. However, those diseases sharing some visual patterns, \emph{smaller inter-class image variation}, are difficult to be recognized. Those images from one class but with disparate visual patterns, \emph{larger intra-class image variation}, such as the flower colors in different growth stage, are also challenging to classify. From the perspective of agriculture, it is inevitable to have smaller inter-class and larger intra-class image variations. Therefore, deep learning methods are expected to mitigate the challenge. In general, those test images with similar image variations as the training images tends to receive high performance. By contrast, deep learning methods are expected to have poor generalization ability, such that models training only with the images with controlled imaging environments will have low performance when it is tested in the images with uncontrolled one \cite{guth2023lab, wu2023laboratory}.

\begin{figure}[htbp]
    \centering
    % \includesvg[inkscapelatex=false, width=1.0\textwidth]{variation.svg}
    \includegraphics[width=1.0\linewidth]{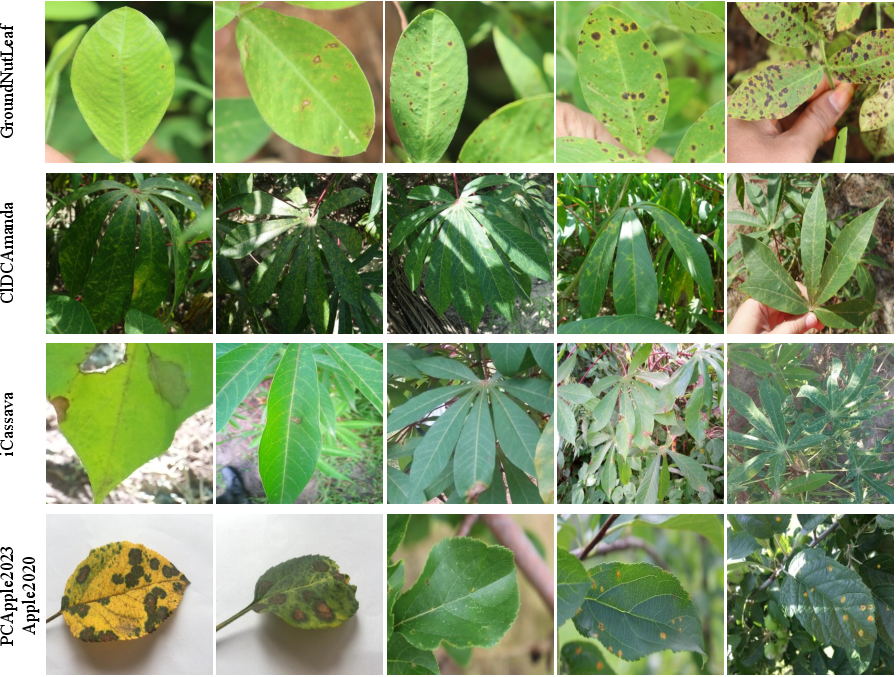}
    \caption{Examples of some image variations from the first to last row: disease stage, illumination, scale, and background. The images are taken from the corresponding datasets. In this paper, image background is grouped into three cases: simple (the first two images), medium (the third and fourth images), and complex (the last image).}
    \label{fig:variation}
\end{figure}

Main variations are summarized from the perspective of forming image variations in Table \ref{tab:variation} and Figure \ref{fig:variation} illustrates some image variations. Some variations are highly related. For example, those images for the plants in the field may have much larger diversity on illumination than the counterparts of greenhouse and laboratory. Similarly, canopies tends to have smaller scales than leaves and fruits. Besides, additional factors may be shared source of multiple variations such that person habits to take pictures result in diversity in scales and viewpoints. We emphasize that we group background as uncontrolled and controlled. For examples, leaves are put on the homogeneous materials such as papers in the laboratories or field. In the field, plant organs of interest can also be moved to have simple background. In contrast, with the uncontrolled background, images are taken without considering the background. Therefore, backgrounds vary significantly and can be controlled when taking pictures for the plants in fields.

\begin{table}[htbp]
    \centering
    \caption{Main variations may cause image variations, partially summarized from \cite{xu2023embracing}. }
    \label{tab:variation}
    \begin{tabularx}{\textwidth}{|l|X|}
        \hline
        Category & Variation \\
        \hline
        \multirow[]{4}{*}{Plant} & Type of plant such as tomato and apple. \\
            & Plant organ including leaf, fruit, stem, flower, and canopy. \\
            & Statue of plant such as florescence and disease such as with early symptom. \\
            & Environment including field, greenhouse, and laboratory. \\
        \hline
        Imaging process & Include illumination, scale, viewpoint, and background. \\
        \hline
    \end{tabularx}
\end{table}

\subsection{Dataset splitting}

In general, three datasets are adopted to develop deep learning models. Training datasets are to train models and validation datasets are taken to choose the optimal set of hyper-parameters such as the architecture of models. After training and validating, test datasets are finally used to check the performance of the trained models. In practical applications, the training and validation datasets are available in the training process whereas the test dataset is not available until users use the trained model. In such case, the test performance is not known and thus model developers can not assess the trained models. An alternative scheme is splitting a hold-out dataset as test before training models, resembling the real test data. Empirical results suggest that splitting results in different performance and thus the splitting strategies should be considered. To be clear, the collected dataset is referred as original one that will be split into three parts, training, validation, and test. The real test data from model users are distinguished from the split test dataset.

The most widely used strategy is \emph{random} splitting that an image in the original dataset is randomly put into one of the three datasets. One of the main issues of this strategy is that multiple images taken for a same observation (a symptom of plant disease in nature) with only slightly differences can be assigned into the training and test dataset, by which the test performance could be overestimated. Moreover, this strategy elusively ignores the generalization challenge \cite{arjovsky2019invariant, bengio2021deep, corso2023holistic, xu2023embracing}, commonly existing in real-world applications, such that the real test data are not in the same distribution of the training dataset.

To mitigate the issues, splitting in \emph{spatial} and \emph{temporal} is appealing. For example, images taken in a place are put into either training or test \cite{beery2022auto}. Similarly, images on the same day or in the same year can be used for only one of the three datasets. In spite of being invariant to the type of plant diseases, the spatial and temporal factors \emph{explicitly} allows that the training and test datasets are in different distributions. Although the strategies introduce new challenges, such as domain shift \cite{xu2023embracing} as suggested in \cite{beery2022auto}, it is worthwhile. In a nutshell, our objective is to achieve the best performance in real test process, rather than neither in the training nor in the split test datasets. For example, a model trained in the collected data from several farms in this year is probably desired to be deployed in different farms for next couple of years. Beyond spatial and temporal, more things can be considered and are encouraged, such the images taken by same \emph{person} who may be with certain habits to take pictures.

\subsection{Annotation strategy}
\label{annotation}
To begin with, we explicitly propose the rules for annotating datasets as follows:
\begin{itemize}
    \item Annotations are difficult, time-consuming, and expensive to obtain.
    \item Annotations from human are with bias and noise.
    \item Images and their annotations should simultaneously satisfy the requirements of deep learning methods and agricultural tasks.
\end{itemize}

% which part is interested such as blurring area?
% 

% existence of annotation
The first rule triggers the first question whether an image is annotated, termed \emph{existence} of annotation. All images are fully labeled, which is the most case in the related public datasets. On the contrary, all images in a collected dataset can be not annotated completely. A more reasonable case is partially with annotations where some images are labeled whereas other images are not, mainly because the images are more easily available in a relative manner. In such scenario, two more factors should be considered: image level, whether an image should be annotated, and class level, how many images should be annotated for a class. We argue that the partially annotated datasets should be raised considering the characteristic of practical applications. Besides, theory also support it as a margin distribution of images can be useful with the learned conditional distribution or joint distribution, between labels and images \cite{bengio2013representation}.

% correct label, noise
Furthermore, bias and noise appear when human make decisions and the magnitude may be underestimated \cite{kahneman2021noise}. For example, the validation dataset of ImageNet \cite{deng2009imagenet}, used to perform image classification of generic objects such as dogs and cats, has around 6\% wrong labels \cite{northcutt2021pervasive}. Compared to this case, plant disease annotation requires more domain knowledge and may be more difficult to be precise. For example, three experts got only 85.9\% accuracy on average when labeling 999 wheat images \cite{long2023classification}. Noisy annotations in the training datasets could result in an unstable training process and inferior test performance whereas the noises in validation datasets may lead to wrong selection of hyper-parameters \cite{patrini2017making}. Deep learning generally assumes that the annotations are correct and tends to obtain better performance if the annotation noise is smaller \cite{patrini2017making}. Based on this observation, making precise annotations is worthwhile yet requires more resources. For example, voting independent decisions from multiple experts tends to be beneficial \cite{kahneman2021noise}. In additional, Polymerase Chain Reaction (PCR) may also contribute \cite{pereira2023enhancing}. We emphasize that we are not trying to say that the bias and noise should be avoided completely but that they should be noticed when annotating and degrade them considering the trade-off, as well as in the model training and validation stages.

The last law highlights the format of annotation, called levels of the annotation. In generic, image classification can be perform in image-level annotation, an image with an label. Multi-label image classification is also possible if an image has multiple plant diseases. However, bounding boxes can point out the location of every instance of plant disease in an image and thus is called instance level annotation. Moreover, pixel-level annotations are desired to the task of segmentation which assign a label for every pixel. For every level of annotation, extra strategies exist, such as the EEP \cite{xu2023embracing}, exclusion that every annotation includes only one specific visual pattern of plant disease, extensiveness that every plant disease in the images should have been annotated, and precision that annotation is expected to be precise for different tasks such as the correct labels and precise location of bounding boxes. Again, we highlight that incompatible formats of annotation become feasible with the concept of weak-supervision \cite{zhou2018brief, xu2023embracing} yet with negative impact on the test performance. Beside, new type of annotations have been emerged and new models may embrace different types of annotation.

\begin{figure}[htbp]
    \centering
    % \includesvg[inkscapelatex=false, width=1.0\textwidth]{object_detection_annotation.svg}
    \includegraphics[width=1.0\linewidth]{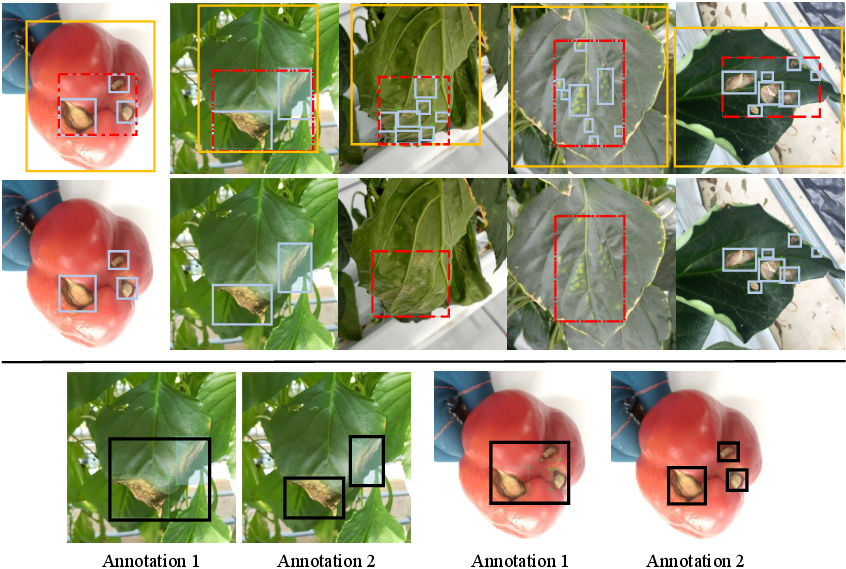}
    \caption{Bounding box annotation strategies in object detection, useful for the localization task. Up first row: three strategies to give bounding box, global (light yellow) that covers one instance such as an instance of fruit or leaf; local (light blue) that covers local areas with dense and intensive symptoms; semi-global (dotted red) that is a trade off of previous two, covering local areas yet allowing a sparse symptoms such as the case in the first image. Up second row: recommended strategy, disease-adaptive, that different diseases may use either local or semi-global strategy. The global one is not recommended because an instance may include more than one type of diseases and may include healthy part. Down: inconsistent annotation when the bounding boxes for the same disease are given in different strategies in a dataset, which may confuse the deep learning model and result in optimizing issues. The picture is adapted from \cite{dong2022data}.}
    \label{fig:object_detection}
\end{figure}

Considering the advantages of localization task using object detection with weak image assumption than image classification and less annotation workload than segmentation \cite{xu2023embracing}, we give more details about it beyond the EEP \cite{xu2023embracing} strategy. To be more specific, how can we give bounding box for \emph{different} diseases in a \emph{consistent} manner, as illustrated in Figure \ref{fig:object_detection}. In general, three independent strategies can be used to have bounding box. First, every instance of fruit or leaf with diseased symptoms is labeled, called global-level. The problem is that an instance may have multiple diseases and thus the corresponding bounding boxes will have diverse labels and include the healthy parts, which may confuse deep learning models or cause challenge to model optimization. Second, every single symptom gets a bounding box and the symptom is assumed to be dense without non-trivial gap, termed local level. In this case, many bounding boxes may exist in an instance, such as the third and fourth images in the first row of Figure \ref{fig:object_detection}, which makes annotation harder and more time-consuming. Besides, different annotators may have diverse definitions about the "dense". Third, semi-level is trade off of previous two, allowing a gap between symptoms, especially for those tiny ones but many. Based on our understanding and experimental results, an adaptive strategy \cite{dong2022data} is recommended that different disease has different levels of annotation. Another elusive issue is the inconsistency mentioned by \href{https://www.google.com/search?q=data+centric+andrew+ng%2C+tomato+annotation&oq=data+&gs_lcrp=EgZjaHJvbWUqCAgAEEUYJxg7MggIABBFGCcYOzIGCAEQRRg7MgYIAhBFGD0yBggDEEUYPDIGCAQQRRg9MgYIBRBFGEEyBggGEEUYQTIGCAcQRRhB0gEINzI5OWowajeoAgCwAgA&sourceid=chrome&ie=UTF-8#fpstate=ive&vld=cid:d80688e6,vid:TU6u_T-s68Y,st:519}{Andrew Ng in a video}. The underlying is that different annotators or even same annotator in different time would use different level of bounding boxes, as compared of annotations 1 and 2 for same image in Figure \ref{fig:object_detection}. This inconsistency in training datasets gives different information to models resulting in unstable learning. In addition, inconsistency in test process may give us an inaccurate evaluation.

% \subsection{Discussion}
% A taxonomy of plant disease recognition dataset is proposed in this section. Although we attempt to give more details as possible, it is difficult to achieve and, to some extent, impossible.

% Although we attempt to give extensive analysis, extra information is encouraged. 

% \subsection{Challenges}

% generalization

% Charactericstic of a standard dataset for deep learning

\section{Public Plant Disease Recognition Datasets}

Based on our preliminary survey, RGB images taken by handled cameras dominate in public plant disease recognition datasets. Other types of datasets are rarely utilized and public. Therefore, we aim to provide a survey on RGB images to recognize plant diseases in this section. We did not do a complete survey that is impossible to some extent and rather focus on those datasets with a relatively higher frequency of utilization or released recently, which suggests the tendency in this field. For a dataset, the following tags are considered:

\begin{itemize}
    \item Dataset name. We will assign a name for the dataset if without a given in the original material. The datasets are default public available. Some partial public datasets are also included.
    \item Crop. If only one crop is included, their name is given. Otherwise, the number of crops is given.
    \item Number of classes. Disease classes and healthy one are included.
    \item Number of images. Only those images with public available annotations are counted.
    \item Image background (BG). We split the image background into three categories as shown in the last row in Figure \ref{fig:variation}. The simple one are taken in the environments of laboratory where the region of interest (RoI) are put on the controlled material. The complex one is taken in the field with complex background. The medium one is also taken in the field but the RoIs may be moved to have a simpler background. Their corresponding abbreviations are sim, med, cmpx.
    \item Machine learning (ML) task and official performance (PE). This paper focuses on three types of machine learning task as discussed before, image classification (clf), object detection (obj), segmentation (seg). A dataset may support more than one task. We only report official performance, either in the original publications or in the leaderboard of official challenges. Otherwise, N.A suggests not available.
\end{itemize}

Table \ref{tab:datasets} summarizes the related public datasets and \href{https://github.com/xml94/PPDRD}{our project} is public in Github with more detailed information. Although we tried to do our best, some beneficial datasets may be not included and thus any new contribution is welcome. One of the main observations is lacking of enough descriptions about objectives and usages. As mentioned above, localization and quantitation analysis are also beneficial but few relevant datasets with relative high-quality are public available. Another observation is that decent performance is achieved in most of the datasets with reported performance. An exception that a model trained in the PlantVillage with simple image background suffers in the FieldPV dataset with different levels of background. Similar situation appears in the FieldPlant dataset. Besides, we found that no datasets are utilized by the majority of research and compared, except the PlantVillage dataset.

\begin{table}[htbp]
    \centering
    \caption{Overview of some public plant disease recognition datasets. Class, image, image BG, ML task, and PE suggest number of classes, number of images, image background, machine learning task, and official performance.}
    \label{tab:datasets}
    % \tiny
    \footnotesize
    % \small
    \begin{tabular}{|l|l|r|r|l|l|}
    \hline
        Dataset name & Crop & Class & Image & Image BG & ML task \& PE \\ \hline
        \href{https://www.kaggle.com/competitions/plant-pathology-2020-fgvc7/overview}{Apple2020} \cite{thapa2020plant} & Apple & 4 & 1,821 & med & clf: 0.984 AUROC \\ \hline
        \href{https://www.kaggle.com/competitions/plant-pathology-2021-fgvc8/data}{Apple2021} & Apple & 6 & 18,632 & med & clf: 0.883 F1 \\ \hline
        \href{https://aistudio.baidu.com/datasetdetail/215559}{PCApple2023} & Apple & 9 & 10,212 & med+sim & clf: N.A \\ \hline
        \href{https://datadryad.org/stash/dataset/doi:10.5061/dryad.41ns1rnj3}{ASDID} \cite{bevers2022soybean} & Soybean & 8 & 9,648 & med+sim & clf: 0.968 Acc \\ \hline
        \href{https://data.mendeley.com/datasets/yy2k5y8mxg/1}{BRACOL} \cite{esgario2020deep} & Coffee & 5 & 1,747 & sim & clf: 0.956 Acc \\ \hline
        \href{https://data.mendeley.com/datasets/c5yvn32dzg/2}{RoCoLe} \cite{parraga2019rocole} & Coffee & 6 & 1,560 & med & clf: N.A \\ \hline
        \href{https://www.kaggle.com/competitions/cassava-disease}{iCassava} \cite{mwebaze2019icassava} & Cassava & 5 & 5,656 & med & clf: 0.939 Acc \\ \hline
        \href{https://www.kaggle.com/competitions/cassava-leaf-disease-classification}{CLDCMakerere} & Cassava & 5 & 21,397 & cmpx+med & clf: 0.913 Acc \\ \hline
        \href{https://scholarsphere.psu.edu/resources/215d1acd-2c1e-440b-a27a-03d212761ef7}{CLDCAmanda} \cite{ramcharan2017deep} & Cassava & 6 & 2,249 & med & clf: 0.930 Acc \\ \hline
        \href{https://data.mendeley.com/datasets/3832tx2cb2/1}{CLDD} & Cassava & 3 & 228 & med & clf: N.A \\ \hline
        \href{https://data.mendeley.com/datasets/y6d3z6f8z9/1}{CDRD} \cite{sultana2023dataset} & Cucumber & 8 & 1,289 & med+sim & clf: N.A \\ \hline
        \href{https://www.kaggle.com/datasets/kareem3egm/cucumber-plant-diseases-dataset}{CucumberNegm} & Cucumber & 2 & 691 & med & clf: N.A \\ \hline
        \href{https://www.kaggle.com/competitions/paddy-disease-classification/overview}{PaddyDoctor} \cite{petchiammal2023paddy} & Rice & 10 & 10,407 & cmpx & clf: 0.990 Acc \\ \hline
        \href{https://drive.google.com/drive/folders/1ewBesJcguriVTX8sRJseCDbXAF_T4akK}{Rice1426} \cite{rahman2020identification} & Rice & 9 & 1,426 & cmpx+med+sim & clf: 0.971 Acc \\ \hline
        \href{https://data.mendeley.com/datasets/fwcj7stb8r/1}{Rice5932} \cite{sethy2020deep} & Rice & 4 & 5,932 & med & clf: 0.984 Acc \\ \hline
        \href{https://www.kaggle.com/datasets/minhhuy2810/rice-diseases-image-dataset}{HuyDoRice} & Rice & 4 & 3,355 & sim & clf: 0.984 Acc \\ \hline
        \href{https://data.mendeley.com/datasets/znsxdctwtt/1}{DhanShomadhan} \cite{hossain2023dhan} & Rice & 5 & 1,106 & cmpx+sim & clf: N.A \\ \hline
        \href{https://www.ebi.ac.uk/biostudies/BioImages/studies/S-BIAD682?query=Septoria}{WheatLong} \cite{long2023classification} & Wheat & 5 & 999 & cmpx & clf: 0.971 Acc \\ \hline
        \href{https://data.mendeley.com/datasets/wgd66f8n6h/1}{WheatLeafDataset} & Wheat & 3 & 407 & med+sim & clf: N.A \\ \hline
        \href{https://data.mendeley.com/datasets/22p2vcbxfk/3}{GroundNutLeaf} \cite{aishwarya2023dataset} & Groundnut & 5 & 3,058 & med & clf: N.A \\ \hline
        \href{https://researchdata.up.ac.za/articles/dataset/Diseases_of_maize_in_the_field/20237613}{MaizeCraze} & Corn & 6 & 2,355 & sim & clf: N.A \\ \hline
        BisqueCorn \href{https://bisque.cyverse.org/client_service/view?resource=https://bisque.cyverse.org/data_service/00-fsRrwb8afr4Q4diBdiWtF9}{1} \href{https://bisque.cyverse.org/client_service/view?resource=https://bisque.cyverse.org/data_service/00-fsRrwb8afr4Q4diBdiWtF9}{2} & Corn & 2 & 1,785 & cmpx & clf: N.A \\ \hline
        \href{https://osf.io/p67rz/}{CornNLB} \cite{wiesner2018image} & Corn & 1 & 18,222 & cmpx & clf: N.A \\ \hline
        \href{https://github.com/AI-Lab-Makerere/ibean}{iBean} & Bean & 3 & 1,296 & med & clf: N.A \\ \hline
        \href{https://data.mendeley.com/datasets/bycbh73438/1}{SoybeanMignoni} \cite{mignoni2022soybean} & Soybean & 3 & 6,410 & cmpx & clf: N.A \\ \hline
        \href{https://data.mendeley.com/datasets/ngdgg79rzb/1}{TaiwanTomato} & Tomato & 6 & 622 & med+sim & clf: N.A \\ \hline
        \href{https://data.mendeley.com/datasets/x84p2g3k6z/1}{GLFD} \cite{rajbongshi2022comprehensive} & Guava & 5 & 527 & sim & clf: N.A \\ \hline
        \href{https://data.mendeley.com/datasets/3f83gxmv57/2}{CitrusRauf} & Citrus & 10 & 759 & sim & clf: N.A \\ \hline
        PlantVillage \cite{hughes2015open} & 14 & 38 & 54,305 & sim & clf: N.A \\ \hline
        \href{https://github.com/PatrickGui/FPDR/tree/master#experimental-data}{FieldPV} \cite{gui2021towards} & 14 & 38 & 665 & med+sim & clf: 0.720 Acc \\ \hline
        \href{https://github.com/pratikkayal/PlantDoc-Dataset}{PlantDocCls} \cite{singh2020plantdoc} & 13 & 27 & 2,598 & cmpx+med+sim & clf: N.A \\ \hline
        \href{https://data.mendeley.com/datasets/hb74ynkjcn/5}{PlantConservation} \cite{chouhan2019data} & 12 & 10 & 4,503 & sim & clf: N.A \\ \hline
        \href{https://data.mendeley.com/datasets/bwh3zbpkpv/1}{CCMT} \cite{mensah2023ccmt} & 4 & 22 & 24,881 & med & clf: N.A \\ \hline
        \href{https://github.com/liuxindazz/PDD271}{PDD271} \cite{liu2021plant} & N.A & 271 & 2,710 & cmpx+med & clf: 0.855 Acc \\ \hline
        \href{https://github.com/pratikkayal/PlantDoc-Dataset}{PlantDocObj} \cite{singh2020plantdoc} & 13 & 27 & 2,598 & cmpx+med+sim & obj: N.A \\ \hline
        \href{https://github.com/kmarif/NZDLPlantDisease-v1}{NZDLPlantDiseaseV1} \cite{saleem2022performance} & 5 & 20 & 3,337 & med & obj: 0.745 mAP \\ \hline
        \href{https://github.com/kmarif/NZDLPlantDisease-v2}{NZDLPlantDiseaseV2} \cite{saleem2022weight} & 8 & 28 & 3,039 & med & obj: 0.932 mAP \\ \hline
        \href{https://universe.roboflow.com/plant-disease-detection/fieldplant}{FieldPlant} \cite{moupojou2023fieldplant} & 4 & 31 & 5,156 & cmpx+med & obj: 0.144 mAP \\ \hline
        \href{https://data.mendeley.com/datasets/3dr9r3w3jn/2}{GrapevineDiseaseMalo} & Grape & 3 & 744 & cmpx & obj: N.A \\ \hline
        \href{https://data.mendeley.com/datasets/3dr9r3w3jn/2}{GrapevineDiseaseMalo} & Grape & 4 & 128 & cmpx & seg: N.A \\ \hline
        \href{https://data.mendeley.com/datasets/yy2k5y8mxg/1}{BRACOL} \cite{esgario2020deep} & Coffee & 2 & 1,560 & sim & seg: N.A \\ \hline
        \href{https://www.scidb.cn/en/detail?dataSetId=0e1f57004db842f99668d82183afd578}{ATLDSD} & Apple & 5 & 1,641 & med+sim & seg: N.A \\ \hline
    \end{tabular}
\end{table}

\section{Future Direction of Plant Disease Recognition Datasets}

\begin{itemize}
    \item Stage one: \emph{verification} where deep learning methods are verified to be useful to recognize plant diseases.
    \item Stage two: \emph{implementation} where deep learning methods are deployed into real-world applications of plant disease recognition with decent performance.
    \item Stage three: \emph{connection} where plant disease recognition using deep learning methods is connected to downstream applications.
\end{itemize}

To probe the future direction, We declare three stages of plant disease recognition using deep learning. The first stage is straightforward and almost finished in recent years. However, the second and third stages are still in their infancy. Currently, few publications mentioned the successful implementation in real-world applications. One of the main reasons comes from the assumptions embraced by deep learning methods, that are generally not hold in real-world applications. From this perspective, existing datasets catered for the assumptions. Therefore, one of the future directions of plant disease recognition is to make datasets violating the assumptions, termed deep learning challenge-oriented dataset. Furthermore, we argue that recognition plant disease is not the final objective and should be connected with the downstream work, arriving the third stage. From such perspective, we argue that another future direction is to make the datasets oriented by downstream applications. Besides, achieving better performance is desired in general, two inspirations are discussed, multi-observation and large-scale datasets. The mentioned content are detailed in the following content with additional discussions.

\subsection{Deep learning challenge-oriented dataset}

Although decent performance is achieved in most datasets, the corresponding trained models may suffer when deploying them in real-world applications. One of the reasons is that the assumptions to achieve good performance when training in the existing datasets are not always held \cite{xu2023embracing}. Violating those assumptions results in challenges to deploy the deep learning models. For example, a model trained in the datasets from several farms is desired to give superior results when deploying in other farms, termed \emph{spatial generalization}. In a similar spirit, a model trained in the datasets collected in a time duration is desired to be decent when deploying in another time duration, termed \emph{temporal generalization}. Additional disease-invariant characteristics are also expected to have no impacts. However, current datasets do not support this kind of verification. Formally, deep learning challenge-oriented datasets are highlighted to test and develop models cater for plant disease recognition. Simultaneously, we argue that datasets should be with meta-data, such as position and time stamp. In the other words, current datasets elusively assume that something in the training and test datasets are shared \cite{yao2023known}. For example, a new plant disease may exist in the test stage and is desired to be classified from the known classes that exist in the training dataset \cite{yao2023known}, the challenge termed open set recognition. Furthermore, we highlight that realizing the assumptions of deep learning models and incorporating them into the dataset collection stage need the cooperation of researchers from agriculture and deep learning field. Please refer to detailed challenges from the datasets in \cite{xu2023embracing}.

\subsection{Application-oriented dataset}
From the perspective of agriculture, recognizing plant diseases may not be the final objective and downstream work may follow. For example, early visual pattern recognition is beneficial to make some remedies to reduce loss. Therefore, collecting such datasets is appealing. Although some papers aims to dedicate this issue, the definition of early disease recognition has no agreement. We contend that such data have two primary characteristics: pattern recognizable and remedies effective. One of the core assumptions embraced by deep learning models is that different plant diseases have their own patterns; otherwise they can not be distinguished. Considering that data modalities have heterogeneous advantages and disadvantages, selecting suitable input modality is essential. On the other hand, statues of diseases cause varying losses and difficulties to take remedies. In an extreme scenario, a plant disease explodes in a farm and plants may all die where recognizing the corresponding disease is not that kind of useful. More application objectives from agriculture field are possible. Although the objectives are from the agriculture, we highlight that trade off exist such as the case in early disease recognition, which requires the cooperation from engineers in deep learning field.

\subsection{Multi-observation dataset}

In general, human experts make superior decisions to recognize plant diseases by looking through multiple observations than single observations. Inspired by this situation, we contend that deep learning methods can also be raised with multi-observation. In a nut shell, multiple observations distribute different information. \textbf{Multi-modal datasets} refer to those datasets with various modalities for the same plant diseases. For example, given a plant disease leaf, various optical images and texts can be made. Besides, datasets can be in \textbf{time-series}, such as taking images for plant diseases in different time. Specially, visual patterns become more clear and easier to be recognized when diseases gradually involve. Time-series datasets may mitigate the challenge of early plant disease recognition. For images data, higher test performance can also be dedicated to \textbf{multi-spatial datasets}, such as taking images in different scales and perspectives. For example, some plant diseases have different patterns in the front and back of leaves.

\subsection{Large-scale dataset}
Large-scale datasets tend to be beneficial for model generalization in many general computer vision tasks and datasets \cite{kaplan2020scaling, zhai2022scaling, xu2023plantclef2023}. Therefore, collecting large-scale datasets for plant disease recognition is appealing and worthwhile although it is time-consuming, difficult, and expensive. One convincing manner is crowdsensing \cite{coletta2020optimal} by which related people in different locations take images and then upload them to a platform. Those images will be annotated by the community. In this way, the collected datasets have enormous variations and thus contribute to model generalization \cite{xu2023embracing}.

\subsection{Discussion}

\textbf{Benchmark} In recent years, plant disease recognition has witnessed a significant improvement \cite{singh2018deep, liujun2021plant, thakur2022trends, salman2023crop, xu2023embracing}, as well as the number of related publications. However, the relative comparison is relative lacking to evaluate different models in diverse applications. One of the main reasons is the shortage of benchmarks, public and widely used high-quality datasets. We argue that this kind of benchmarks will facilitate the community and speed up the deployment of deep learning methods in the real-world applications of plant disease recognition.

\textbf{Meta-data} is the documentary to describe data instances from different perspectives, usually with tags. Most of current relevant public datasets have only the types of plant disease. Other types of information are expected to be beneficial, such as the spatial and temporary tags. Those dataset with meta-data can be used for different applications by making new datasets.

\textbf{Analysis of datasets} In general, different applications have heterogeneous difficulties and challenges. Datasets show the faces of applications and therefore, analysis of datasets are essential to understand the applications and further to achieve a better performance. However, few datasets have corresponding analysis and one of the expected directions is automatic analysis, such as for intra- and inter-class image variations. Furthermore, dataset analysis can be used in an iterate way to make high-quality datasets.

\textbf{Beyond plant disease recognition} Recognizing plant disease is just one of the fundamental requirements to have decent crop yields. This objective may be further facilitated through incorporating recognition and more things. For example, plants may be infected by specific diseases or virus in some conditions where finding the correlated factors are beneficial to prevent the plants from those diseases.

% Analyzing the image variations is beneficial, a kind of prior of human knowledge.
% https://huggingface.co/docs/datasets/v1.12.0/dataset_card.html
% \emph{}
% \subsubsection{Circle of creating a dataset}
% dataset creating is always continuous, not finished.
% \subsubsection{Data privacy and safety}
% beyond recognition: forecast and incidence analysis
% annotation with text.

\section{Conclusion}

Although plant disease recognition has witnessed a significant improvement in recent years, its applications in real-world still surfer and one of the reasons stems from the datasets. Making a high-quality plant disease recognition dataset requires superior understandings in both agriculture field and deep learning. This study aimed to provide the perspective for the creation of plant disease recognition datasets. To depict them clearly, we first provided a systematic taxonomy for related datasets. We specially emphasize dataset splitting and annotation strategies that are few discussed in the literature and elusively suggest challenges in real-world applications. Further, RGB images are observed as the dominate input modality and an extensive summarization was provided. Finally, four kinds of dataset as future directions are described: deep learning challenge-oriented, application-oriented, multi-observation, and large-scale, with an additional discussion. We believe that this paper will enhance the understandings about plant disease recognition and their datasets and is beneficial to make new datasets with the objective, implementing deep learning methods in real-world applications. To facilitate the field, \href{https://github.com/xml94/PPDRD}{our project} is public available and new datasets are encouraged.

%% The Appendices part is started with the command \appendix;
%% appendix sections are then done as normal sections
% \appendix

%% If you have bibdatabase file and want bibtex to generate the
%% bibitems, please use
%%
 \bibliographystyle{elsarticle-num} 
 \bibliography{cas-refs}

%% else use the following coding to input the bibitems directly in the
%% TeX file.

% \begin{thebibliography}{00}

% %% \bibitem{label}
% %% Text of bibliographic item

% \bibitem{}

% \end{thebibliography}
\end{document}